\documentclass[lettersize,journal]{IEEEtran}
\usepackage{amsmath,amsfonts}
\usepackage{array}
\usepackage[caption=false,font=normalsize,labelfont=sf,textfont=sf]{subfig}
\usepackage{textcomp}
\usepackage{stfloats}
\usepackage{url}
\usepackage{verbatim}
\usepackage{graphicx}
\usepackage{cite}

\usepackage{algorithm}
\usepackage{algorithmicx}
\usepackage{algpseudocode}
\usepackage{array}
\usepackage[caption=false,font=normalsize,labelfont=sf,textfont=sf]{subfig}
\usepackage{textcomp}
\usepackage{stfloats}
\usepackage{url}
\usepackage{verbatim}
\usepackage{graphicx}
\usepackage{xcolor}
\usepackage{colortbl}
\usepackage{amsmath}

\usepackage[utf8]{inputenc} % allow utf-8 input
\usepackage[T1]{fontenc}    % use 8-bit T1 fonts
\usepackage{hyperref}       % hyperlinks
\usepackage{url}            % simple URL typesetting
\usepackage{booktabs}       % professional-quality tables
\usepackage{amsfonts}       % blackboard math symbols
\usepackage{nicefrac}       % compact symbols for 1/2, etc.
\usepackage{microtype}      % microtypography
\usepackage{xcolor}         % colors

\usepackage{microtype}
\usepackage{graphicx}
\usepackage{subcaption}
\usepackage{booktabs} % for professional tables
\usepackage{url}
\usepackage{placeins}

\usepackage{amssymb}
\usepackage{mathtools}
\usepackage{amsthm}
\usepackage{bbm}
\usepackage{multirow}
\usepackage{makecell}
\usepackage{float}

\usepackage[capitalize,noabbrev]{cleveref}

\begin{document}

\title{BlockGaussian: Efficient Large-Scale Scene Novel View Synthesis via Adaptive Block-Based Gaussian Splatting}

% ******************************author names****************************** 
\author{Yongchang Wu, Zipeng Qi, Zhenwei Shi and Zhengxia Zou$^\star$

\thanks{The work was supported by the National Key Research and Development
Program of China under Grant 2022ZD0160401, the National Natural Science Foundation of China under Grant 62125102, 62471014, and U24B20177, the Beijing Natural Science Foundation under Grant JL23005, and the Fundamental Research Funds for the Central Universities. \emph{(Corresponding author: Zhengxia Zou (e-mail: zhengxiazou@buaa.edu.cn))}}
\thanks{Yongchang Wu, Zipeng Qi, Zhengxia Zou, and Zhenwei Shi are with the Department of Aerospace Intelligent Science and Technology, School of Astronautics, Beihang University, Beijing 100191, China; the State Key Laboratory of Virtual Reality Technology and Systems, Beihang University; and Shanghai Artificial Intelligence Laboratory, Shanghai 200232, China.}
}



\twocolumn[{
\renewcommand\twocolumn[1][]{#1}%
\maketitle
\begin{center}
    \centering
    \captionsetup{type=figure}
    \includegraphics[width=\textwidth]{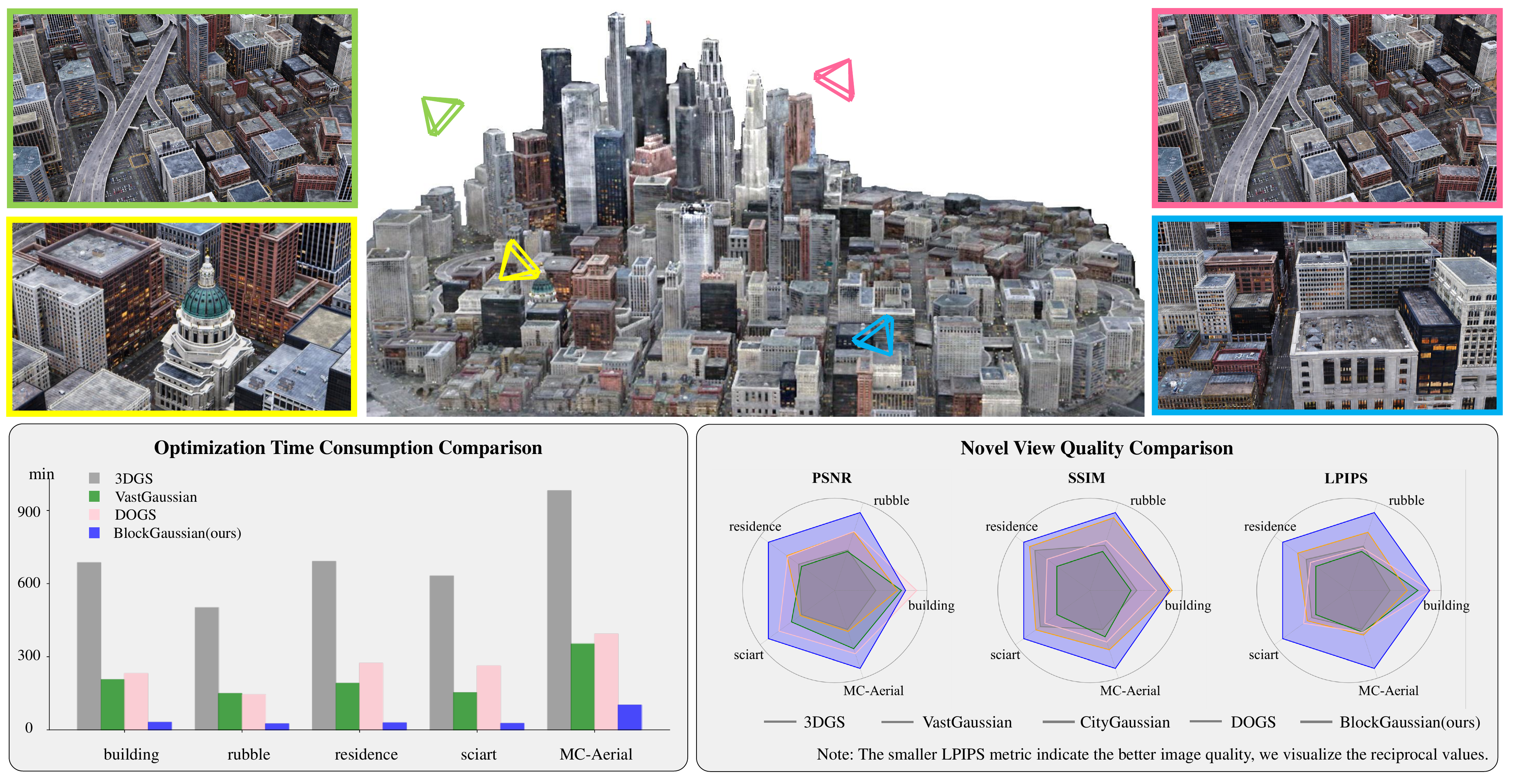}
    \captionof{figure}{BlockGaussian reconstructs city-scale scenes from massive multi-view images and enables high-quality novel view synthesis from arbitrary viewpoints, as illustrated in the surrounding images. Compared to existing methods, our approach reduces reconstruction time from hours to minutes while achieving superior rendering quality in most scenes.}
    \vspace{1.5em}
    \label{fig:teaser}
\end{center}
}]

\begin{abstract}
The recent advancements in 3D Gaussian Splatting (3DGS) have demonstrated remarkable potential in novel view synthesis tasks. The divide-and-conquer paradigm has enabled large-scale scene reconstruction, but significant challenges remain in scene partitioning, optimization, and merging processes. This paper introduces BlockGaussian, a novel framework incorporating a content-aware scene partition strategy and visibility-aware block optimization to achieve efficient and high-quality large-scale scene reconstruction.
Specifically, our approach considers the content-complexity variation across different regions and balances computational load during scene partitioning, enabling efficient scene reconstruction.  To tackle the supervision mismatch issue during independent block optimization, we introduce auxiliary points during individual block optimization to align the ground-truth supervision, which enhances the reconstruction quality. Furthermore, we propose a pseudo-view geometry constraint that effectively mitigates rendering degradation caused by airspace floaters during block merging.
Extensive experiments on large-scale scenes demonstrate that our approach achieves state-of-the-art performance in both reconstruction efficiency and rendering quality, with a 5× speedup in optimization and an average PSNR improvement of 1.21 dB on multiple benchmarks.
Notably, BlockGaussian significantly reduces computational requirements, enabling large-scale scene reconstruction on a single 24GB VRAM device. The project page is available at \url{https://github.com/SunshineWYC/BlockGaussian}.
\end{abstract}


\section{Introduction}
\IEEEPARstart{L}{arge}-scale scene high-fidelity and real-time novel view synthesis is essential for many applications, including autonomous driving\cite{yang2020surfelgan, yan2024street, 10891659}, virtual reality\cite{cho2019novel, wang2025can}, remote sensing photogrammetry\cite{wu2022remote, xu2024mega}, and embodied intelligence. Recently, prominent novel view synthesis approaches have fallen into two main categories: Neural Radiance Fields (NeRF)-based methods\cite{mildenhall2021nerf, barron2021mip, barron2022mip, gao2022nerf} and Gaussian Splatting-based techniques\cite{kerbl20233d, huang20242d, chen2024survey}. Neural Radiance Fields (NeRF)\cite{mildenhall2021nerf}, due to their capability in high-fidelity rendering through implicit representation, have been extended to large-scale scene reconstruction tasks\cite{turki2022mega,zhenxing2022switch,tancik2022block}. Although Block-NeRF\cite{tancik2022block} achieved large-scale reconstruction of San Francisco neighborhoods, the scene representation using MLP networks as the smallest unit lacks flexibility and struggles with slow rendering speeds. 3D Gaussian Splatting\cite{kerbl20233d}, as an alternative, demonstrates more significant potential, particularly with its fast rendering speed. Explicit point clouds scene representation makes it more scalable for large-scale scenes\cite{kerbl2024hierarchical, lin2024vastgaussian, liu2024citygaussian, chen2025dogs}.

The divide-and-conquer paradigm\cite{turki2022mega, liu2024citygaussian, lin2024vastgaussian} has become the mainstream of large-scale scene novel view synthesis constrained by hardware VRAM resources. By dividing the scene into subregions, the reconstruction speed has been significantly improved in multi-GPU parallel. This paradigm consists of three key stages: scene partitioning, individual block optimization, and fusion of block reconstruction results. These stages exhibit a strictly sequential dependency, meaning the input of each subsequent stage entirely relies on the output of the preceding phase. The quality of the final reconstruction result is contingent upon the effectiveness of each stage. Although existing research methods have established a baseline, challenges remain in these steps, including imbalanced reconstruction complexity across blocks, supervision mismatch in block-wise optimization, and quality degradation in fusion results.

\begin{figure*}[t]
    \centering
    \includegraphics[width=\textwidth]{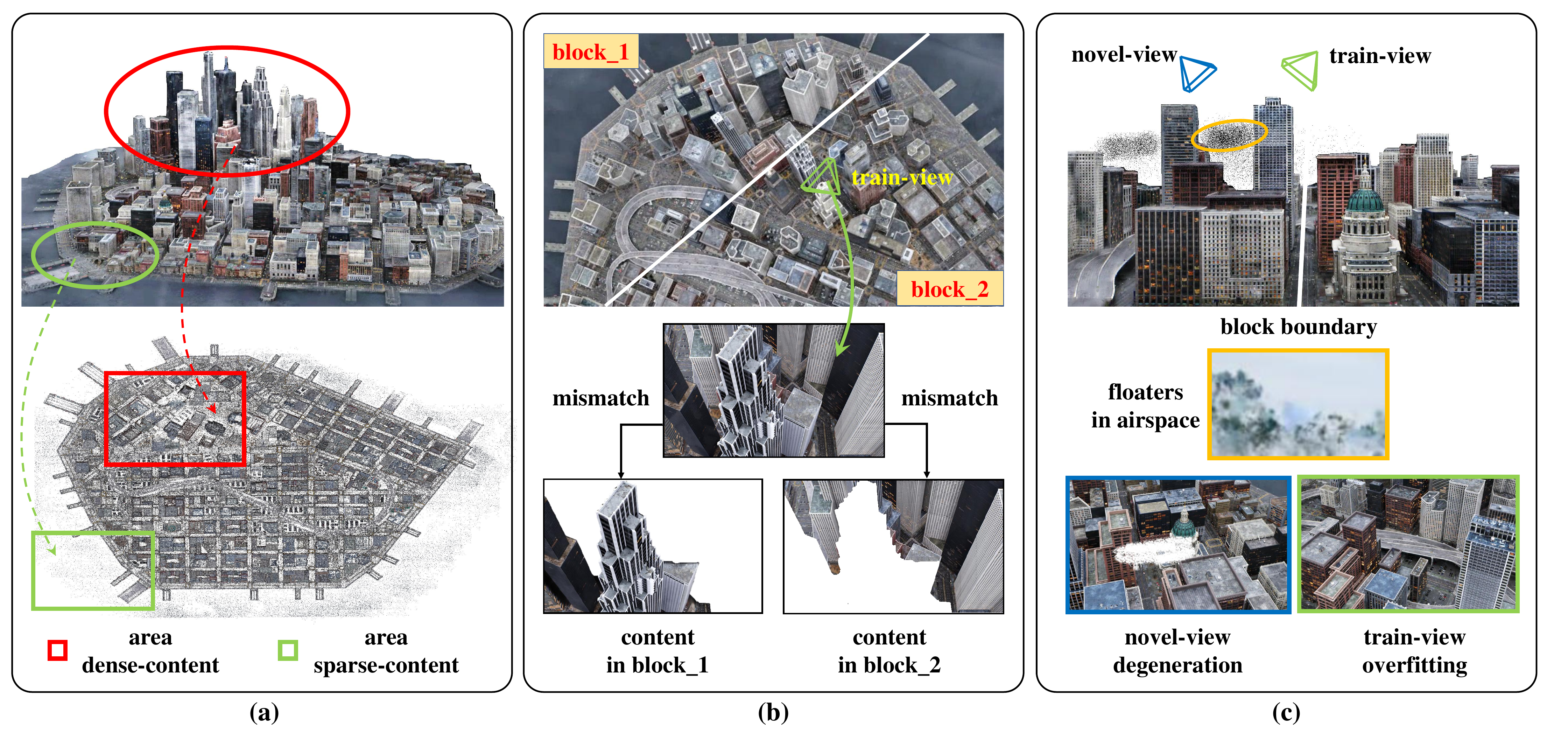}
    \caption{Existing challenges in large-scale scene novel view synthesis task under divide-and-conquer paradigm. a) Imbalanced reconstruction complexity across blocks: The intensity of content in different scene regions exhibits significant differences. Areas with dense content require finer subdivision granularity to ensure reconstruction fidelity, while sparser-content regions benefit from coarser partitioning to enhance computational efficiency. b) Supervision mismatch in block-wise optimization: The content of a training view may be divided into multiple blocks after scene partitioning. Due to visibility constraints, the entire training view image does not match the ideal supervision when optimizing the individual block. c) Quality degradation in fusion results: Floater in airspace is an important reason for the quality degradation of fusion results. Since each block is optimized individually, these floaters fit well in the training perspective but degrade the quality of the synthesized novel views, especially in the boundary region.}
    \label{fig:challenges}
\end{figure*}

The imbalanced reconstruction complexity across blocks arises from unreasonable scene partitioning, which reduces the efficiency of large-scale scene reconstruction, especially on multi-GPU devices. As illustrated in Fig.\ref{fig:challenges} (a), evenly dividing the scene into grids ignores the content disparity of different regions. When partitioning the scene, two critical factors must be considered: the granularity of block division and the computational loads across blocks. The former requires attention to the complexity of different scene regions, with higher granularity for areas of interest or greater complexity. The latter aims to balance the computational loads across blocks, thereby reducing the training time of the entire scene in multi-GPU. VastGaussian\cite{lin2024vastgaussian} introduced a progressive data partitioning strategy, which divides the scene based on camera positions. DOGS\cite{chen2025dogs} improved VastGaussian's partitioning method by proposing a recursive method to balance computational loads across blocks. However, scene partitioning based on camera positions is limited by the spatial distribution of cameras and struggles to generalize to scenes with more complex viewpoint distributions. CityGaussian\cite{liu2024citygaussian} firstly trains a coarse Gaussian as the scene prior, using it to partition the scene into grids. Nevertheless, this partitioning method requires pre-training a coarse Gaussian model, which does not fully decouple scene scale from the optimization process.

The supervision mismatch in block-wise optimization leads to artifacts within individual blocks, degrading scene reconstruction quality. NeRF/Gaussian-based scene representations employ an end-to-end optimization pipeline, where the parameters of the scene representation are optimized by constructing an objective function that compares ground-truth images with rendered images. However, under the divide-and-conquer paradigm for large-scale scene reconstruction, the lack of a entire scene representation leads to visibility problems during individual block optimization. As illustrated in Fig.\ref{fig:challenges} (b), after scene partitioning, the content of a training view may be distributed across multiple blocks, indicating that a single block only corresponds to a portion of the image scope. When reconstructing the concerned block, a mismatch arises between the rendered image and the training view during loss calculation. This mismatch stems from two factors: a) the image rendering process from the concerned block ignores occlusion relationships between blocks; b) it is difficult to calculate the accurate boundary with under-optimized block representations. Noisy supervision confuses the gradients of Gaussian parameters during end-to-end optimization, leading to degradation reconstruction results.

% \begin{figure}
%     \centering
%     \includegraphics[width=0.47\textwidth]{images/mismatch.pdf}
%     \caption{Supervision mismatch arises when optimizing individual blocks. After scene partitioning, the content of a training view may be distributed across multiple blocks, indicating that a single block only corresponds to a portion of the image content. The regions of the image that do not appear in \textit{block\_1} can lead to degraded artifacts if directly using the entire image to supervise the reconstruction of \textit{block\_1}. Resolving this issue is essential for ensuring accurate and artifact-free reconstruction in the divide-and-conquer paradigm.}
%     \label{fig:mismatch}
% \end{figure}

Seamless scene fusion avoiding quality degradation is another critical challenge in large-scale scene reconstruction. Optimizing individual blocks tends to produce floaters in the airspace due to the lack of accurate geometric supervision, leading to degenerate solutions. This significantly degrades the rendering quality after block fusion. As shown in Fig.\ref{fig:challenges} (c), floaters in airspace may fit the training-views well, but cause artifacts in novel views, especially at block boundaries. Therefore, adequate airspace supervision is vital for the scene training process. VastGaussian\cite{lin2024vastgaussian} tries to solve this problem by introducing more training views and designs an airspace-aware visibility calculation method that selects viewpoints based on the proportion of the projected polygon of the block's boundary. However, this approach has two limitations: it ignores occlusion relationships between blocks, and the selected viewpoints tend to introduce additional regions outside the block, raising a contradiction between viewpoint selection and adequate supervision.

To address these challenges, we propose BlockGaussian, a new framework for large-scale novel view synthesis. In the scene division stage, we propose a spatial-based scene partitioning method termed Content-Aware Scene Partition, which dynamically and finely divides the scene based on sparse point cloud output from the prior Structure from Motion\cite{schonberger2016structure} process while comprehensively considering the computational loads across multiple blocks.
To relieve the supervision mismatch issue during individual block reconstruction, we remodel the single-block optimization problem and propose a visibility-aware optimization algorithm. During the optimization process, auxiliary point clouds are introduced to adaptively represent the invisible regions of the training view to relieve the supervision mismatch issue. The experimental results demonstrate the effectiveness of the auxiliary point cloud. 
For airspace supervision, given the complexity of occlusion relationships in the scene, directly selecting viewpoints which can provide adequate airspace supervision for the current block is challenging. Unlike VastGaussian\cite{lin2024vastgaussian}, we design a Pseudo-View Geometry Constraint to supervise airspace without introducing regions outside the concerned block. Specifically, we perturb training camera poses to generate pseudo viewpoints. With the rendered depth maps, we warp the ground truth images from the original viewpoints and compute the loss corresponding to the images rendered from the pseudo viewpoints. This constraint significantly improves block fusion quality, especially for interactive rendering.

Experimental results demonstrate that our proposed method effectively addresses the challenges in large-scale scene reconstruction. As shown in Fig.\ref{fig:teaser}, in terms of both reconstruction quality and speed, BlockGaussian achieves state-of-the-art (SOTA) performance across multiple scenes, with a 5× speedup in optimization and an average PSNR improvement of 1.21 dB. Regarding hardware requirements, BlockGaussian can be executed sequentially on a single 24GB VRAM GPU or parallel across multiple GPUs. Furthermore, our method exhibits strong generalization capabilities, performing well in aerial-view scenes and street-view scenes.
Our contributions are summarized as follows:

1) We propose BlockGaussian for large-scale scene novel view synthesis, with a spatial-based scene partitioning paradigm that dynamically balances the granularity of block division and the computational loads across blocks.

2) We remodel the training process of individual blocks and propose a visibility-aware block optimization method by introducing auxiliary point clouds to address the mismatch between rendered images and supervised viewpoints.

3) We introduce a novel pseudo-view geometry constraint to supervise the airspace. This constraint can effectively mitigate rendering quality degradation caused by airspace floaters during block fusion, ensuring seamless and high-quality scene fusion.

\section{Related Work\label{related_work}}
\subsection{Novel View Synthesis}
Novel view synthesis (NVS) is a fundamental problem in computer vision and graphics, aiming to synthesize photorealistic images of a scene from viewpoints that were not captured initially. Early image-based rendering techniques, such as light field rendering \cite{levoy2023light} and view morphing \cite{seitz1996view}, use collections of images to synthesize novel views by interpolating between captured viewpoints. These methods often assume dense scene sampling, limiting their practicality in real-world scenarios. Depth-based methods, such as 3D warping \cite{mcmillan1995head}, use depth maps to project pixels from source images to the target viewpoint. While effective, these approaches are sensitive to depth estimation errors and often produce artifacts in regions with occlusions or complex geometry.

The advent of deep learning has revolutionized NVS, enabling data-driven approaches that learn to synthesize novel views directly from images without explicit 3D reconstruction. DeepStereo\cite{flynn2016deepstereo} uses a deep network to predict novel views from a sparse set of input images by learning to interpolate between them. Similarly, Multi-plane images(MPI) \cite{zhou2018stereo} represents scenes as layered depth images and use neural networks to refine and synthesize novel views.

Differentiable rendering techniques have significantly propelled progress in novel view synthesis methods. Neural Radiance Fields \cite{mildenhall2021nerf} stands as a landmark approach, representing a scene as a continuous volumetric function parameterized by a neural network. Subsequent works have improved upon vanilla NeRF in various aspects, including reconstruction and rendering efficiency\cite{fridovich2022plenoxels, chen2022tensorf, muller2022instant}, anti-aliasing\cite{barron2021mip, barron2022mip}, sparse input\cite{yuan2022neural}, appearance consistency\cite{martin2021nerf, 10416701}, and scene generalizability\cite{chen2021mvsnerf, wang2021ibrnet, nguyen2023cascaded}.
Gaussian Splatting\cite{kerbl20233d}, achieving real-time novel view synthesis through efficient rasterization, has emerged as another milestone in novel view synthesis. SparseGS\cite{xiong2024sparsegs} focuses on few-shot novel view synthesis and introduces depth-regularization to remove floater artifacts. LightGaussian\cite{fan2024lightgaussian} and Compact3d\cite{navaneet2023compact3d} manage to reduce the model size and remove redundant 3D Gaussians through exhaustive quantization. The explicit representation inherent in 3D Gaussian Splatting facilitates its application in downstream tasks, such as autonomous driving \cite{yan2024street}, human body representation\cite{kocabas2024hugs}, navigation\cite{keetha2024splatam, matsuki2024gaussian, 10870413} and so on.

\subsection{Large Scale Scene Reconstruction}
Significant progress has been made in large-scale scene reconstruction in recent years. Traditional scene reconstruction pipelines\cite{agarwal2011building, schonberger2016structure} typically involve several sequential steps: feature extraction and matching, camera parameter estimation, dense reconstruction, meshing, and texture mapping, which collectively recover the geometry and appearance of the scene. Structure from Motion (SfM) encompasses feature extraction, matching, camera parameter estimation, outputting camera parameters and sparse point clouds of the scene. Due to its stability and robustness, SfM based on feature points and bundle adjustment remains the mainstream framework for pose estimation and sparse reconstruction. Traditional appearance reconstruction pipelines take the camera parameters as input, generate dense depth maps through multi-view stereo (MVS) methods\cite{bleyer2011patchmatch, galliani2015massively, 9076298, 8765737, 9863705}, and then produce a mesh-based scene representation via meshing\cite{6180172} and texture mapping\cite{7780814, 9184935}. In recent years, with the development of differentiable rendering techniques, end-to-end optimization-based reconstruction methods, such as Neural Radiance Fields\cite{mildenhall2021nerf} and 3D Gaussian Splatting\cite{kerbl20233d}, have achieved superior reconstruction results compared to traditional step-by-step approaches.

For large-scale scenes, the divide-and-conquer strategy is a widely adopted solution for handling massive datasets. This approach divides the scene into grids, optimizes each grid separately, and then merges the reconstruction results to obtain a complete scene representation. NeRF-based methods have successfully reconstructed street and aerial views of large-scale scenes. For instance, Block-NeRF\cite{tancik2022block} reconstructs San Francisco neighborhoods from 2.8 million street-view images and accounts for transient objects and appearance variations by modifying the underlying NeRF architecture. Mega-NeRF\cite{turki2022mega} introduces a sparse and spatially aware network structure to represent aerial scenes and makes valuable attempts at interactive rendering. Switch-NeRF\cite{zhenxing2022switch} designs a learnable scene decomposition based on sparse large-scale NeRF representations. Grid-NeRF\cite{xu2023grid} integrates MLP-based NeRF with feature grids to encode local and global scene information. Although its two-branch design achieves high visual fidelity in rendering, it remains constrained by the slow training and rendering speeds inherent to NeRF.
On the other hand, due to significant advantages in rendering speed, gaussian-based methods are increasingly being explored for large-scale scene applications. Hierarchy-GS\cite{kerbl2024hierarchical} proposes a hierarchical representation and optimizes chunk parameters in parallel, leveraging the divide-and-conquer strategy. Scaffold-GS \cite{lu2024scaffold} combines explicit and implicit representations, achieving a more compact scene representation while maintaining high-quality view synthesis. Octree-GS\cite{ren2024octree} introduce Level-of-detail (LOD) to 3D Gaussian Splatting, using a novel octree structure to organize anchor Gaussians hierarchically to achieve real-time rendering.

Concurrent works with our method include VastGaussian\cite{lin2024vastgaussian}, CityGaussian\cite{liu2024citygaussian}, and DOGS\cite{chen2025dogs}. These three methods all adopt a paradigm of scene partitioning, viewpoint assignment, parallel optimization, and scene fusion to reconstruct large-scale scenes. VastGaussian designs a progressive data partitioning strategy to divide the scene and allocate training views. DOGS refines the scene partitioning process with a recursive approach, aiming to split the scene into blocks with a more balanced distribution of cameras. However, both methods rely on the camera position distribution for block partitioning, neglecting the misalignment between the scene content distribution and the camera distribution. This limitation poses challenges for downstream tasks, such as dynamic map loading. In CityGaussian, a global coarse Gaussian model is first trained to guide scene partitioning and viewpoint allocation, which becomes difficult to implement under limited computational resources. Unlike VastGaussian and DOGS, which partition the scene based on camera position, our method proposes a spatial-based block partitioning strategy. This approach ensures a more flexible and adaptive division of the scene and balances the computational loads between blocks. Additionally, we remodel the individual block optimization problem, propose a visibility-aware optimization algorithm, and design a pseudo-view geometry constraint during optimization. These innovations effectively mitigate rendering inconsistencies and improve the quality of individual block reconstructions, facilitating seamless block merging and improving the fidelity of synthesized views.

\section{Preliminary\label{Preliminary}}
This section briefly introduces the vanilla 3D Gaussian Splatting on which our BlockGaussian is based. 3D Gaussian Splatting utilizes discrete Gaussian primitives in 3D space, denoted as $\mathcal{G} = \{\mathbf{G}_k \}$, where each gaussian primitive $\mathbf{G}_k$ consists of learnable attributes, including position $\mathbf{x}_k$, rotation $\mathbf{R}_k$, opacity $o_k$, scales $\mathbf{s}_k$, and spherical harmonics(SH)\cite{ramamoorthi2001efficient} coefficients $\mathbf{f}_k$. During the rendering process, each 3D Gaussian primitive is projected onto the image plane as a 2D Gaussian. Volume rendering\cite{martin2021nerf, kerbl20233d} is then performed to compute the final RGB values for each pixel. The rendering process can be formulated as follows:
\begin{equation}
\mathbf{C} = \sum_{i=1}^{N} \alpha_i \mathbf{c_i} \prod_{j=1}^{i-1} (1 - \alpha_j)
\label{eqn:rendering}
\end{equation}
where $\mathbf{C}$ denotes the pixel color, while $\mathbf{c}_i$ represents the RGB color of the Gaussian primitive computed based on spherical harmonics (SH) features $\mathbf{f}_i$. $\alpha$ refers to the transparency weight derived from the projected 2D Gaussian covariance and the Gaussian opacity $o_k$.

Similarly, the depth map ${D}$ can be computed pixel-by-pixel following the alpha blending process\cite{xiong2024sparsegs}. Here, $d_i$ represents the depth value of the gaussian primitive's center point in the camera space.
\begin{equation}
D(u, v) = \sum_{i=1}^{N} \alpha_i d_i \prod_{j=1}^{i-1} (1 - \alpha_j)
\end{equation}

The reconstruction and optimization of the scene commence with the known viewpoints $\mathcal{V} = \{ (I_{i}^{\text{gt}}, \mathbf{R}_i, \mathbf{t}_i) \} $ and the sparse point cloud. Here, $I_{i}^{\text{gt}}$, $\mathbf{R}_i$, and $\textbf{t}_i$ denote the ground truth image, camera orientation, and camera position for the $i$-th viewpoint, respectively. The camera poses $(\mathbf{R}_i, \mathbf{t}_i)$ along with the sparse point cloud $\mathbf{P}$, are estimated through the Structure from Motion (SfM) process. For each training view, the rendered image is computed following $I_{i}^{r} = \mathcal{R}(\mathbf{R}_i, \mathbf{t}_i, \mathcal{G})$. The parameters of the 3D Gaussian are optimized by minimizing the loss between the rendered image and the ground truth image.
\begin{equation}
\mathcal{L}(I_i^{\text{gt}}, I_i^\text{r}) = (1 - \lambda) \mathcal{L}_1(I_i^\text{r}, I_i^{\text{gt}}) + \lambda \mathcal{L}_{\text{SSIM}}(I_i^\text{r}, I_i^{\text{gt}})
\end{equation}

$\lambda$ represents a weighting hyperparameter. To enhance the reconstruction quality of fine details, the densification process is conducted concurrently along with optimization, which supplements the scene representation with additional optimizable variables based on gradient information.

\begin{figure*}[t]
    \centering
    \includegraphics[width=\textwidth]{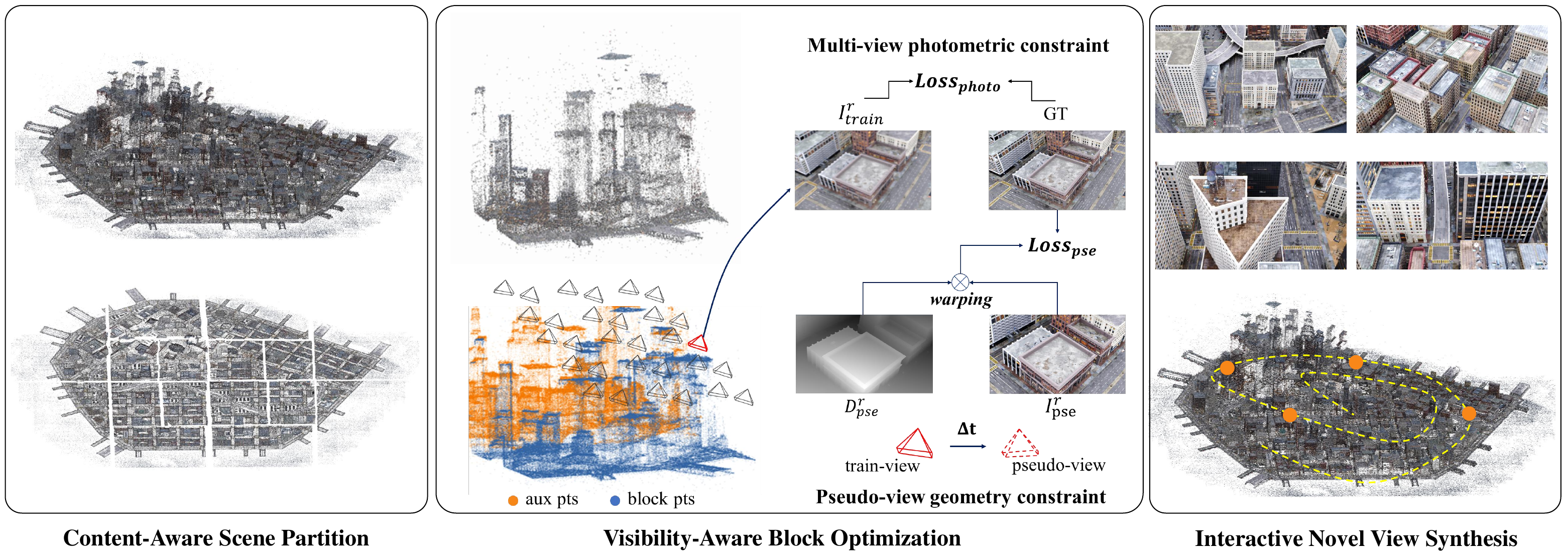}
    \caption{Overview of our proposed method. We first divide the entire scene and allocates viewpoints with Content-Aware Scene Partition, which jointly considering the complexity of scene content and the computational load distribution across blocks. Subsequently, we optimize each block independently, which is executable either sequentially on a single GPU or in parallel across multiple GPUs. During block optimization, we introduce auxiliary point clouds (aux pts) to address supervision mismatch issues. Pseudo-View Geometry Constraint is conducted to supervise airspace regions and mitigate floater artifacts. Finally, the optimized results from all blocks are integrated to construct a comprehensive Gaussian Representation of the entire scene, enabling interactive novel view synthesis.}
    \label{fig:overall}
\end{figure*}

\section{Method\label{method}}
Large-scale scenes present significant challenges due to the extensive area and massive amounts of data. We adopt the divide-and-conquer paradigm similar to previous work\cite{turki2022mega, lin2024vastgaussian, liu2024citygaussian, chen2025dogs}. The overview of our method is shown in Fig.\ref{fig:overall}. Given a collection of captured images, we first calculate the camera poses and sparse point cloud for each viewpoint with Structure from Motion.
Then, we iteratively partition the scene into blocks and assign supervised views to each block with our Context-Aware Scene Partition module, as detailed in Section \ref{partition}. Subsequently, blocks are trained separately under Visibility-Aware Block Optimization, as discussed in Section \ref{block_optimization}. Section \ref{pseudo-constraint} elaborates the Pseudo-View Geometry Constraint provides airspace supervision. Finally, we seamlessly integrate all the blocks to obtain a unified scene representation in Section \ref{merging}.

\subsection{Content-Aware Scene Partition\label{partition}}
Scene partitioning and view assignment are critical steps in reconstructing large-scale scenes. When partitioning the scene, it is essential to balance the trade-off between the granularity of the blocks and the speed of parallel optimization. Intuitively, higher granularity in block partitioning can improve reconstruction quality, but it often leads to slow reconstruction speed. Conversely, lower granularity reduces the time cost of the reconstruction process but at the expense of decreased reconstruction precision. Therefore, during scene partitioning and view assignment, two primary objectives must be satisfied:

1) Adaptive partitioning based on scene complexity: The partitioning should adapt to the complexity of the spatial scene structure, applying different granularity levels to regions with varying importance and complexity.

2) Balanced computational load: The partitioning should ensure an even distribution of computational load across blocks, which is crucial for minimizing the time required for multi-GPU scene reconstruction.

The density distribution of the sparse point cloud in a scene can serve as an estimator of the scene's content complexity. Based on this assumption, we recursively partition the scene into multiple blocks. Specifically, given the sparse point cloud $\mathbf{P}_s$ of the scene, we estimate the normal direction of the ground plane using the Manhattan world assumption\cite{coughlan1999manhattan} and align it with the y-axis. The sparse point clouds are projected onto the x-z plane, and a bounding rectangle is manually defined as the region of interest (RoI) for reconstruction. Subsequently, we partition the reconstruction area into multiple blocks in a binary tree structure. Given the maximum depth $M$ of the binary tree and the maximum number of points $N_b^t$ in a leaf node, the initial RoI region of the scene is taken as the root node for recursive partitioning. if the current node's depth $d<M$ and the number of contained point clouds $N_b > N_b^t$, it is bisected along the longest edge of the corresponding block to generate two child nodes. Otherwise, the partitioning terminates, marking the node as a leaf node. This process is iteratively executed until all nodes meet the termination condition, thereby achieving a spatially adaptive scene partitioning.

The view assignment process aims to select appropriate supervisory training views for each block. We score the relevancy between each training view and the blocks. The number of visible key points $N_v$ for each training view can be obtained from the Structure from Motion results. For each block, we can count the number of points $N_b$ within the boundary. Training views with a ratio $N_b^v / N_v$ greater than thresh(0.3) are selected as supervised views for the current block. The detailed scene partitioning and view assignment procedures are outlined in Algorithm \ref{alg:adaptive_scene_partitioning}.

% \begin{algorithm}[t]
% \caption{Adaptive Scene Partitioning and View Assignment}
% \label{alg:adaptive_scene_partitioning}
% \begin{algorithmic}[1]
% \REQUIRE Scene ROI bounding box \( B_s \); Max tree depth \( M \); Block Point Num Threshold \( N_b^t \); Scene Sparse Pointcloud \( \mathbf{P}_s \); View assignment ratio threshold \( \text{ratio}^t \).
% \ENSURE Leaf blocks' boxes \( B \) and assigned view.

% \STATE \textbf{Function} AssignViewForBlock(\( B, P_s \))
% \STATE \( V_b \leftarrow \{ \} \) \COMMENT{Initialize assigned view list \( V_b \).}
% \STATE \( P_b \leftarrow \) select points in block \( B \) from \( \mathbf{P}_s \).
% \FOR {each view \( V \) in train views}
%     \STATE \( N_b^v \leftarrow \) count \( V \) visible points in \( \mathbf{P}_b \).
%     \STATE \( N_v \leftarrow \) the number of points visible in \( V \).
%     \IF{\( N_b^v / N_v < \text{ratio}^t \)}
%         \STATE \( V_b = V_b \cup \{V\} \).
%     \ENDIF
% \ENDFOR
% \RETURN \( V_b, N_b \).
% \STATE \textbf{End Function}
% \State
% \STATE \textbf{Function} PARTITION(\( B_c \), depth)
% \STATE \( V_b, N_b \leftarrow \) AssignViewForBlock(\( B_c, \mathbf{P}_s \)).
% \IF{\( N_b > N_b^t \) AND depth \( < M \)}
%     \STATE \( B_{c1}, B_{c2} \leftarrow \) split block \( B_c \) along longer edge.
%     \STATE PARTITION(\( B_{c1} \), depth + 1).
%     \STATE PARTITION(\( B_{c2} \), depth + 1).
% \ELSE
%     \STATE \( B \leftarrow B \cup \{B_c\} \).
% \ENDIF
% \STATE \textbf{End Function}
% \State
% \STATE \textbf{Execution:} Initialize with PARTITION(\( B_s, 0 \)).
% \end{algorithmic}
% \end{algorithm}

\begin{algorithm}[t]
\caption{Scene Partitioning and View Assignment}
\label{alg:adaptive_scene_partitioning}
\begin{algorithmic}[1]
\Require Scene ROI bounding box \( B_s \); Max tree depth \( M \); Block Point Num Threshold \( N_b^t \); Scene Sparse Pointcloud \( P_s \); View assignment ratio threshold \( \text{ratio}^t \).
\Ensure Leaf blocks' boxes \( B \) and assigned view.

\Function{AssignViewForBlock}{$B, P_s$}
    \State $V_b \gets \{ \}$
    \State $P_b \gets$ select points in block $B$ from $P_s$.
    \For{each view $V$ in train views}
        \State $N_b^v \gets$ count $V$ visible points in $P_b$.
        \State $N_v \gets$ the number of points visible in $V$.
        \If{$N_b^v / N_v < \text{ratio}^t$}
            \State $V_b \gets V_b \cup \{V\}$.
        \EndIf
    \EndFor
    \State \Return $V_b, N_b$.
\EndFunction

\Statex

\Function{Partition}{$B_c$, $d$}
    \State $V_b, N_b \gets$ \Call{AssignViewForBlock}{$B_c, P_s$}.
    \If{$N_b > N_b^t$ \textbf{and} $d < M$}
        \State $B_{c1}, B_{c2} \gets$ split $B_c$ along longer edge.
        \State \Call{Partition}{$B_{c1}$, $d + 1$}.
        \State \Call{Partition}{$B_{c2}$, $d + 1$}.
    \Else
        \State $B \gets B \cup \{B_c\}$.
    \EndIf
\EndFunction

\Statex

\State \textbf{Execution:} Initialize with \Call{Partition}{$B_s, 0$}.
\end{algorithmic}
\end{algorithm}

\subsection{Visibility-Aware Block Optimization\label{block_optimization}}Thanks to well-designed scene partition strategy, the optimization between blocks is completely independent and can be trained in parallel on multiple GPUs. For single block optimization, the Gaussian primitives within the block can be represented as \( \mathcal{G}_b = \{\textbf{G}_b^n\} \), termed \textbf{block Gaussians}. The supervised views associated with the current block are denoted as \(\{I_i^\text{gt}\}\), and the corresponding camera poses are represented as \(\{\mathbf{R}_i, \mathbf{t}_i\}\). Rendering images \( I_i^b = \mathcal{R}(\mathbf{R}_i, \mathbf{t}_i, \mathcal{G}_b) \) only based on the Gaussians within the current block often fails to cover the entire region of training views. This limitation introduces erroneous supervision when computing the loss with the ground truth image \( I_i^{\text{gt}} \).
To address this issue, we introduce auxiliary Gaussian primitives \( \mathcal{G}_a = \{\textbf{G}_a^m\} \), termed \textbf{auxiliary Gaussians}, which model the scene regions outside the current block for the supervised views. Given the camera pose \((\mathbf{R}_i, \mathbf{t}_i)\) of a supervised view, the image rendering process is now expressed as 
\begin{equation}
I_i^\text{r} = \mathcal{R}(\mathbf{R}_i, \mathbf{t}_i, \mathcal{G}_b, \mathcal{G}_a)
\end{equation}

The initialization of block Gaussians follows the same procedure as in vanilla 3D Gaussian Splatting. Specifically, points that locate in the spatial bounds of the block are initialized as $\mathcal{G}_b$. Auxiliary Gaussians $\mathcal{G}_a$ are initialized from the sparse point clouds associated with the supervised views of the current block but out of the block range.

For each view, the photometric loss is the same as that of 3DGS, as mentioned in Section \ref{Preliminary}. In addition, we use depth maps as priors during optimization following the prior works\cite{xiong2024sparsegs, chung2024depth}. Depth maps $D^{\text{e}}$ are estimated by DepthAnythingV2\cite{depth_anything_v2} and calculate absolute error with rendered depths $D^\text{r}$ in inverse space after scale alignment.
\begin{equation}
\mathcal{L}_{\text{depth}}(D^{\text{e}},D^\text{r}) = \mathcal{L}_{1}(\frac{1}{D^{\text{e}}}, \frac{1}{D^\text{r}})
\end{equation}

Given $k$ supervised views for a block, the objective function for the optimization process can be formulated as follows:
\begin{equation}
\underset{\mathcal{G}_a, \mathcal{G}_b}{\mathbf{argmin}} \sum_{i=1}^{k} (\mathcal{L}(I_i^{\text{gt}}, I_i^\text{r}) + \mathcal{L}_{\text{depth}}(D^{\text{e}}_i, D^\text{r}_i))
\end{equation}

It is evident that the optimization of auxiliary Gaussians $\mathcal{G}_a$ suffers from insufficient supervision. Directly applying the same optimization paradigm as in 3D Gaussian Splatting (3DGS) would lead to the degradation of auxiliary Gaussians, which in turn adversely affects the optimization of the primary focus $\mathcal{G}_b$.

To mitigate this issue, we introduce a mini-batch optimization strategy to enhance the stability of the optimization process. Given a collection of supervised perspectives, mini-batch optimization is employed to increase the stability of the gradients. We accumulate the gradients from several training views to update the properties of the 3D representation ${\mathcal{G}_b, \mathcal{G}_a}$. During densification, we only densify the Gaussian points in the block, which effectively reduces the redundancy in optimization process.

\subsection{Pseudo-view Geometry Constraint\label{pseudo-constraint}}
\begin{figure}[t]
    \centering
    \includegraphics[width=0.5\textwidth]{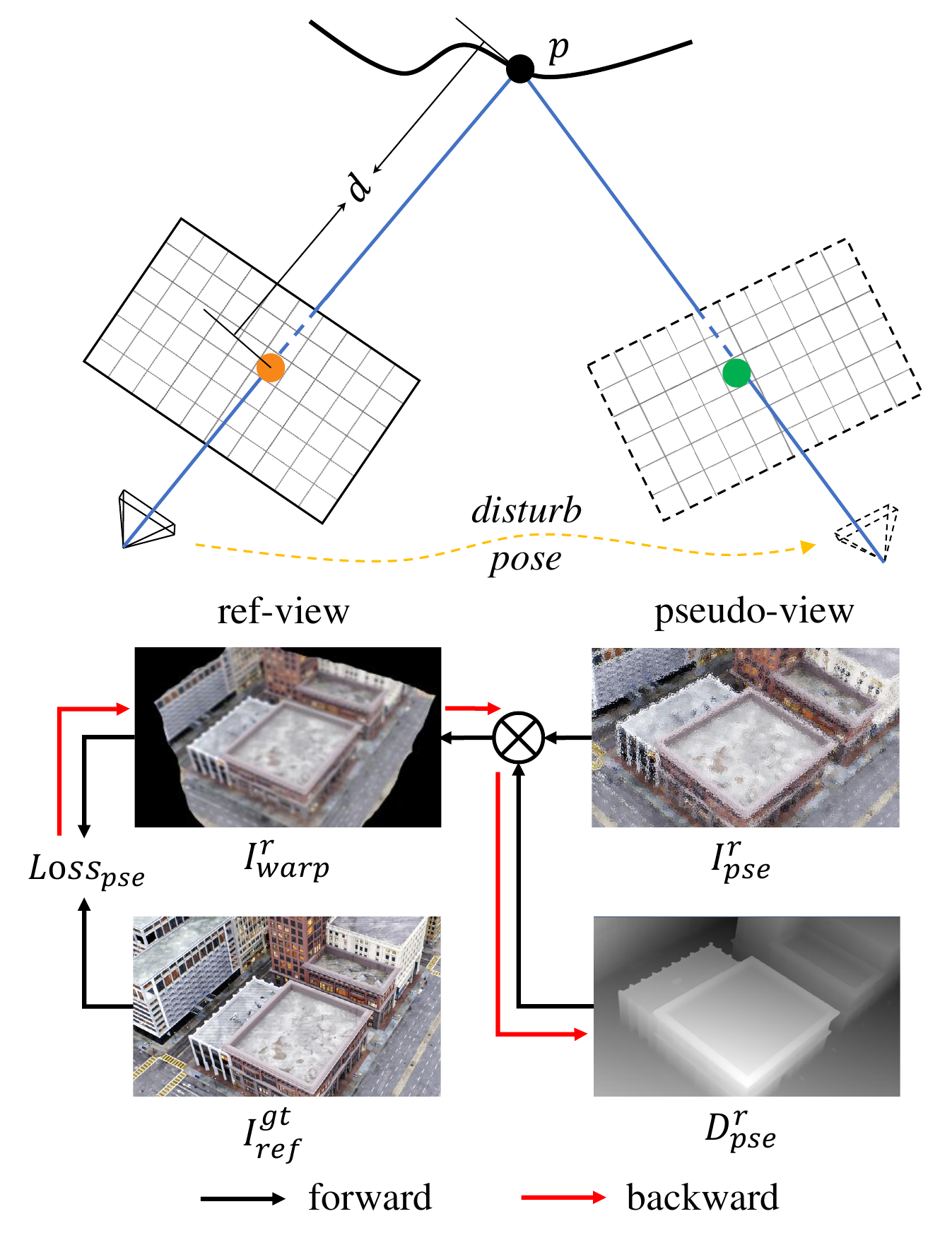}
    \caption{Illustration of the Pseudo-View Geometry Constraint. Typically, artifacts in the airspace can fit RGB images well with inaccurate depth. To address this, we impose constraints on depth to suppress floaters generated in the airspace. For each training view, we generate a pseudo-view by applying slight perturbations to the camera pose. Then we warp the pseudo-view rendered image $I_{\text{pse}}^{\text{r}}$ utilizing rendered depth map $D_{\text{pse}}^{\text{r}}$ to train-view $I_{\text{warp}}^{\text{r}}$. The loss calculated between $I_{\text{warp}}^{\text{r}}$ and train-view ground-truth $I_{\text{ref}}^{\text{gt}}$ provides depth supervision.}
    \label{fig:flowchart_pseudoloss}
\end{figure}

As mentioned in VastGaussian\cite{lin2024vastgaussian}, supervising the airspace during the end-to-end optimization process is crucial. We observe that floaters in the airspace cause the degradation of rendered image quality after scene fusion. To address this issue, we propose a pseudo-view geometry constraint loss to effectively supervise the airspace without introducing additional views to the concerned block.

The process is illustrated in Fig.\ref{fig:flowchart_pseudoloss}. Specifically, for a training view denoted as the reference view (ref-view), whose camera parameters are $(\mathbf{K}_{\text{ref}}, \mathbf{R}_{\text{ref}}, \mathbf{t}_{\text{ref}})$ and whose rendered depth is denoted as $D_{\text{ref}}^{\text{r}}$, we perturb the pose of the ref-view to obtain the camera pose of the pseudo-view. To finely control the magnitude of the perturbation, the disturbance is calculated in disparity space. Given a hyperparameter $\Delta {p}$  representing the disparity perturbation in the horizontal direction, the positional disturbance $\Delta \mathbf{t}$ is computed as
\begin{equation}
\Delta \mathbf{t} = [\frac{median(D_{\text{ref}}^{\text{r}}) \cdot \Delta {p}}{f}, 0, 0]^{T},
\end{equation}
where $median(D_{\text{ref}}^{r})$ refers to the median value of $D_{\text{ref}}^{r}$, $f$ represents the x-axis focal length of ref-view camera. The camera parameters of the pseudo-view are then expressed as:
\begin{equation}
(\mathbf{K}_{\text{pse}}, \mathbf{R}_{\text{pse}}, \mathbf{t}_{\text{pse}}) = (\mathbf{K}_{\text{ref}}, \mathbf{R}_{\text{ref}}, \mathbf{t}_{\text{ref}}+\Delta \mathbf{t})
\end{equation}

We follow the process below to warp the image $I_{\text{pse}}^{\text{r}}$ from the pseudo-view to the ref-view perspective based on the rendered depth map $D_{\text{pse}}^{\text{r}}$. For a pixel in the pseudo-view rendered image $I_{\text{pse}}^{\text{r}}(u_{\text{pse}}, v_{\text{pse}})$, its corresponding depth value $z_{\text{pse}}=D_{\text{pse}}^{\text{r}}(u_{\text{pse}}, v_{\text{pse}})$ is used to compute its projected position in the ref-view. We firstly restore the position of each pixel in pseudo-view camera space with the following equation.
\begin{equation}
\begin{bmatrix} 
x_c^{\text{pse}} \\ 
y_c^{\text{pse}} \\ 
z_c^{\text{pse}}
\end{bmatrix}
=
\mathbf{K}_{\text{pse}}^{-1}
\begin{bmatrix} 
u_{\text{pse}} \\ 
v_{\text{pse}} \\ 
1 
\end{bmatrix} 
\cdot z_{\text{pse}}
\end{equation}

Next, these points are reprojected into the world coordinate system.
\begin{equation}
\begin{bmatrix} 
x_w \\ 
y_w \\ 
z_w
\end{bmatrix}
=
\mathbf{R}_{\text{pse}}^{-1}
\begin{bmatrix} 
x_c^{\text{pse}} \\ 
y_c^{\text{pse}} \\ 
z_c^{\text{pse}}
\end{bmatrix}
- \mathbf{R}_{\text{pse}}^{-1} \cdot \mathbf{t_{\text{pse}}}
\end{equation}

The obtained world coordinates are then transformed into the ref-view camera space using the extrinsic parameters of the reference camera.
\begin{equation}
\begin{bmatrix} 
x_c^{\text{ref}} \\ 
y_c^{\text{ref}} \\ 
z_c^{\text{ref}}
\end{bmatrix}
=
\mathbf{R}_{\text{ref}}
\begin{bmatrix} 
x_w \\ 
y_w \\ 
z_w
\end{bmatrix}
+ \mathbf{t}_{\text{ref}}
\end{equation}

Subsequently, we obtain the corresponding pixel coordinates in the reference-view image space by applying the intrinsic camera transformation.
\begin{equation}
z_{\text{ref}} \cdot
\begin{bmatrix} 
u_{\text{ref}} \\ 
v_{\text{ref}} \\ 
1 
\end{bmatrix} 
=
\mathbf{K}_{\text{ref}}
\begin{bmatrix} 
x_c^{\text{ref}} \\ 
y_c^{\text{ref}} \\ 
z_c^{\text{ref}}
\end{bmatrix}
\end{equation}

By mapping each pixel position individually, we obtain the warped image $I_{\text{warp}}^{\text{r}}$ and the corresponding validity mask $M$. The pseudo-view geometry loss $\mathcal{L}_{\text{pse}}$ is then formulated as follows:
\begin{equation}
\mathcal{L}_{\text{pse}} = M \cdot \mathcal{L}_{1}(I_{\text{ref}}^{\text{gt}}, I_{\text{warp}}^{\text{r}})
\end{equation}

The Pseudo-View Geometry Constraint achieves indirect supervision of the rendered depth, which significantly helps in removing floaters in airspace.

\subsection{Scene Merging and Rendering\label{merging}}
Once all block optimization processes are completed, we merge the block reconstruction results to obtain the entire scene representation. Thanks to the well-designed block optimization process and the pseudo-view geometry constraint, we can directly merge the scene after cropping the auxiliary Gaussians $\mathcal{G}_a$. When rendering the novel view, BlockGaussian follows the same differentiable rendering pipeline as the original 3D Gaussian Splatting framework\cite{kerbl20233d}. Given the target camera pose and intrinsic parameters, the scene representation—composed of all blocks' Gaussian primitives—is projected onto the image plane. These Gaussians are then alpha-blended in a depth-ordered manner to synthesize the view following the Eqn.\ref{eqn:rendering}.

\section{Experiments\label{experiments}}

\subsection{Experiments Setup}
\textbf{Datasets.} We conducted comprehensive evaluations of our proposed methods on three benchmark datasets: Mill19\cite{turki2022mega}, UrbanScene3D\cite{UrbanScene3D}, and MatrixCity\cite{li2023matrixcity}. The Mill19 and UrbanScene3D datasets comprise aerial imagery captured through real-world drones, with each scene containing thousands of high-resolution images. We maintained consistent dataset partitioning with Mega-NeRF in the training and testing phases. To facilitate fair comparison across all experiments, we uniformly applied 4× downsampling to each image following previous approaches\cite{lin2024vastgaussian, liu2024citygaussian}.

\textbf{Metrics.} To quantitatively evaluate the quality of novel view synthesis, we employed three widely recognized metrics: Peak Signal-to-Noise Ratio (PSNR), Structural Similarity Index Measure (SSIM), and Learned Perceptual Image Patch Similarity (LPIPS)\cite{zhang2018unreasonable}. Considering the inherent photometric variations in scene imagery data, we implemented a color correction consistent with VastGaussian to the rendered images for metric computation.
To quantitatively evaluate the efficiency, we report the optimization time consumption, allocated VRAM and the number of Gaussian points of each scene.

\textbf{Compared Methods.} We conducted extensive comparative experiments against state-of-the-art large-scale scene reconstruction methods, which can be categorized into NeRF-based approaches and 3D Gaussian Splatting based methods. The NeRF-based baselines encompass Mega-NeRF\cite{turki2022mega} and Switch-NeRF\cite{zhenxing2022switch}, while the 3DGS-based methods include VastGaussian\cite{lin2024vastgaussian}, CityGaussian\cite{liu2024citygaussian}, DOGS\cite{chen2025dogs}, and modified 3DGS\cite{kerbl20233d}. It is noteworthy that for CityGaussian, the coarse-stage training proved infeasible within 24GB VRAM constraints for certain scenes. Thus, we directly adopted the metrics reported in their original publication. Additionally, it should be noted that DOGS employs a 6× downsampling strategy during image preprocessing, which may bring advantages in quantitative metric evaluation. For the efficiency evaluation, we implement the algorithms on 8 RTX4090 GPUs platform. Notably, cause different methods may have different number of blocks due to the scene partition strategy, we report the total reconstruction time consumption of each scene, ignoring the case that the number of blocks is less than 8.

\textbf{Implementation Details.} Thanks to the complete independence of individual block optimization, the training process can be efficiently parallelized across multiple GPUs or sequentially executed block-by-block on a single GPU. In our proposed method, each block is trained up to 40000 / 60000 iterations (BlockGaussian-40K / BlockGaussian-60K), with densification performed every 200 iterations. During optimization, Pseudo-View Geometry Constraint is conducted from 10k iteration, with loss weight gradually increases logarithmically from 0.1 to 1.0, and the depth regularization weight gradually decreases from 1.0 to 0.1.

\subsection{Comparison with Other Methods}

\begin{table*}[t]
\renewcommand{\arraystretch}{1.3}
\centering
\caption{Quantitative comparison of novel view synthesis results on Mill19\cite{turki2022mega} and UrbanScene3D\cite{UrbanScene3D} dataset. The  best, the second best, and the third best results are highlighted in \colorbox{red!30}{red}, \colorbox{orange!50}{orange} and \colorbox{yellow!50}{yellow}.}
\resizebox{\textwidth}{!}{
\small
\begin{tabular}{cccccccccccccccc}
\hline
\multirow{2}{*}{\textbf{Scenes}} & \multicolumn{3}{c}{\textbf{building}} & & \multicolumn{3}{c}{\textbf{rubble}} & & \multicolumn{3}{c}{\textbf{residence}} &  & \multicolumn{3}{c}{\textbf{sci-art}} \\ 
& PSNR & SSIM & LPIPS & & PSNR & SSIM & LPIPS & & PSNR & SSIM & LPIPS & & PSNR & SSIM & LPIPS \\ \hline
Mega-NeRF & 20.92 & 0.547 & 0.454 & & 24.06 & 0.553 & 0.508 & & 22.08 & 0.628 & 0.401 & & \cellcolor{yellow!50}25.60  & 0.770 & 0.312 \\
Switch-NeRF & 21.54 & 0.579 & 0.397 & & 24.31 & 0.562 & 0.478 & & \cellcolor{yellow!50}22.57 & 0.654 & 0.352 & & \cellcolor{red!30}26.51 & 0.795  & 0.271 \\ \hline
3DGS  & 20.23 & 0.735  & 0.289 & & 25.24 & 0.755 & 0.253 & & 21.21 & 0.791 & 0.232 & & 21.21 & 0.821 & 0.245 \\
VastGaussian & \cellcolor{yellow!50}21.80  & 0.728 & 0.225 & & 25.20  & 0.742 & 0.264 & & 21.01 & 0.699 & 0.261 & & 22.64 & 0.761  & 0.261 \\
CityGaussian & 21.55 & \cellcolor{red!30}0.778 & 0.246 & & 25.77 & \cellcolor{yellow!50}0.813 & \cellcolor{yellow!50}0.228 & & 22.00 & \cellcolor{yellow!50}0.813 & \cellcolor{yellow!50}0.211 & & 21.39 & \cellcolor{yellow!50}0.837 & 0.230  \\
DOGS & \cellcolor{red!30}22.73 & 0.759 & \cellcolor{red!30}0.204 & & \cellcolor{yellow!50}25.78 & 0.765 & 0.257 & & 21.94 & 0.74  & 0.244 & & 24.42 & 0.804  & \cellcolor{yellow!50}0.219 \\ \hline
\textbf{BlockGaussian-40K} & 21.72 & \cellcolor{yellow!50}0.762 & \cellcolor{yellow!50}0.222 & & \cellcolor{orange!50}26.18 & \cellcolor{orange!50}0.816 & \cellcolor{orange!50}0.213 & & \cellcolor{orange!50}22.63  & \cellcolor{orange!50}0.821 & \cellcolor{orange!50}0.196 & & 24.69 & \cellcolor{orange!50}0.848 & \cellcolor{orange!50}0.208 \\
\textbf{BlockGaussian-60K} & \cellcolor{orange!50}22.05 & \cellcolor{orange!50}0.775 & \cellcolor{orange!50}0.206 & & \cellcolor{red!30}26.33 & \cellcolor{red!30}0.824 & \cellcolor{red!30}0.200 & & \cellcolor{red!30}23.25 & \cellcolor{red!30}0.838 & \cellcolor{red!30}0.182 & & \cellcolor{orange!50}25.91 & \cellcolor{red!30}0.881  & \cellcolor{red!30}0.171 \\ \hline
\end{tabular}
}
\label{tab:comparison_u3d}
\end{table*}

\begin{table*}[t]
\renewcommand{\arraystretch}{1.3}
\centering
\caption{Quantitative comparison of novel view synthesis results on Mill19\cite{turki2022mega} and UrbanScene3D\cite{UrbanScene3D} dataset. We present the optimization time OptTime (hh:mm), the number of final points ($10^6$) and the allocated VRAM (GB) during evaluation.}
\resizebox{\textwidth}{!}{
\begin{tabular}{ccccccccccccc}
\hline
\multirow{2}{*}{\textbf{Scenes}} & \multicolumn{3}{c}{\textbf{building}} & \multicolumn{3}{c}{\textbf{rubble}} & \multicolumn{3}{c}{\textbf{residence}} & \multicolumn{3}{c}{\textbf{sci-art}} \\
& OptTime & Points & VRAM & OptTime  & Points & VRAM & OptTime & Points & VRAM & OptTime & Points & VRAM \\ \hline
Mega-NeRF & 19:49 & - & 5.84 & 30:48 & - & 5.88 & 27:20 & - & 5.99 & 27:39 & - & 5.97 \\
Switch-NeRF & 24:46 & - & 5.84 & 38:30 & - & 5.87 & 35:11 & - & 5.94 & 34:34 & - & 5.92 \\ \hline
3DGS & 11:26 & 5.71 & 3.82 & 08:22 & 3.97 & 2.95 & 11:31 & 5.86 & 3.78 & 10:32 & 4.77 & 3.65 \\
VastGaussian & 03:26 & 5.6 & 3.07 & 02:30 & 4.71 & 2.74 & 03:12 & 6.26 & 3.67 & 02:33 & 4.21 & 3.54 \\
CityGaussian & - & 13.30 & 8.80 & - & 9.60 & 6.55 & - & 10.80 & 7.89 & - & 5.37 & 3.49 \\  
DOGS & 03:51 & 6.89 & 3.39 & 02:25 & 4.74 & 2.54 & 04:33 & 7.64 & 6.11 & 04:23 & 5.67 & 3.53 \\ \hline
\textbf{BlockGaussian-40K} & 00:32 & 13.6 & 9.10 & 00:25 & 10.43 & 6.80 & 00:29 & 11.29 & 8.30 & 00:27 & 4.58 & 3.35 \\
\textbf{BlockGaussian-60K} & 01:09 & 17.96 & 11.12 & 00:52 & 12.33 & 7.76 & 01:01 & 12.94 & 10.05 & 00:51 & 5.40 & 3.60 \\ \hline
\end{tabular}
}
\label{tab:comparison_u3d_speed}
\end{table*}

\begin{table*}[t]
\renewcommand{\arraystretch}{1.3}
\centering
\caption{Quantitative comparison on MatrixCity\cite{li2023matrixcity}. The best results are highlighted in \textbf{bold}.}
\small
\begin{tabular}{ccccccccccc}
\hline
\multirow{2}{*}{\textbf{Scenes}} & \multicolumn{5}{c}{\textbf{MatrixCity-Aerial}} & \multicolumn{5}{c}{\textbf{MatrixCity-Street}} \\
& PSNR & SSIM & LPIPS & OptTime & Points & PSNR & SSIM & LPIPS & OptTime & Points\\ \hline
3DGS & 27.83 & 0.821 & 0.229 & 16:21 & 11.5 & 20.92 & 0.655 & 0.624 & 10:22 & 3.56 \\
VastGaussian & 28.33 & 0.835 & 0.22 & 05:53 & 12.5 & - & - & - & - & - \\
CityGaussian & 27.46 & 0.865 & 0.204 & - & 23.7 & - & - & - & - & - \\
DOGS & 28.58 & 0.847 & 0.219 & 06:34 & 10.3 & 21.61 & 0.652 & 0.649 & 02:33 & 2.37 \\
\textbf{BlockGaussian} & \textbf{29.32} & \textbf{0.908} & \textbf{0.112} & \textbf{01:42} & 36.9 & \textbf{25.48} & \textbf{0.821} & \textbf{0.272} & \textbf{00:53} & 5.99 \\ \hline
\end{tabular}
\label{tab:comparison_mc}
\end{table*}

\textbf{Reonstruction Quality.} We present the average PSNR, SSIM, and LPIPS metrics of BlockGaussian across multiple scenes from Mill19, UrbanScene3D, and MatrixCity datasets in Tab.\ref{tab:comparison_u3d} and Tab\ref{tab:comparison_mc}. Compared to existing methods, BlockGaussian-40K achieves comparable performance after 40k training iterations and BlockGaussian-60K outperforms existing methods in most scenes, particularly regarding SSIM and LPIPS metrics, indicating that the synthesized novel views exhibit superior perceptual details.

\begin{figure*}[t]
    \centering
    \includegraphics[width=\textwidth]{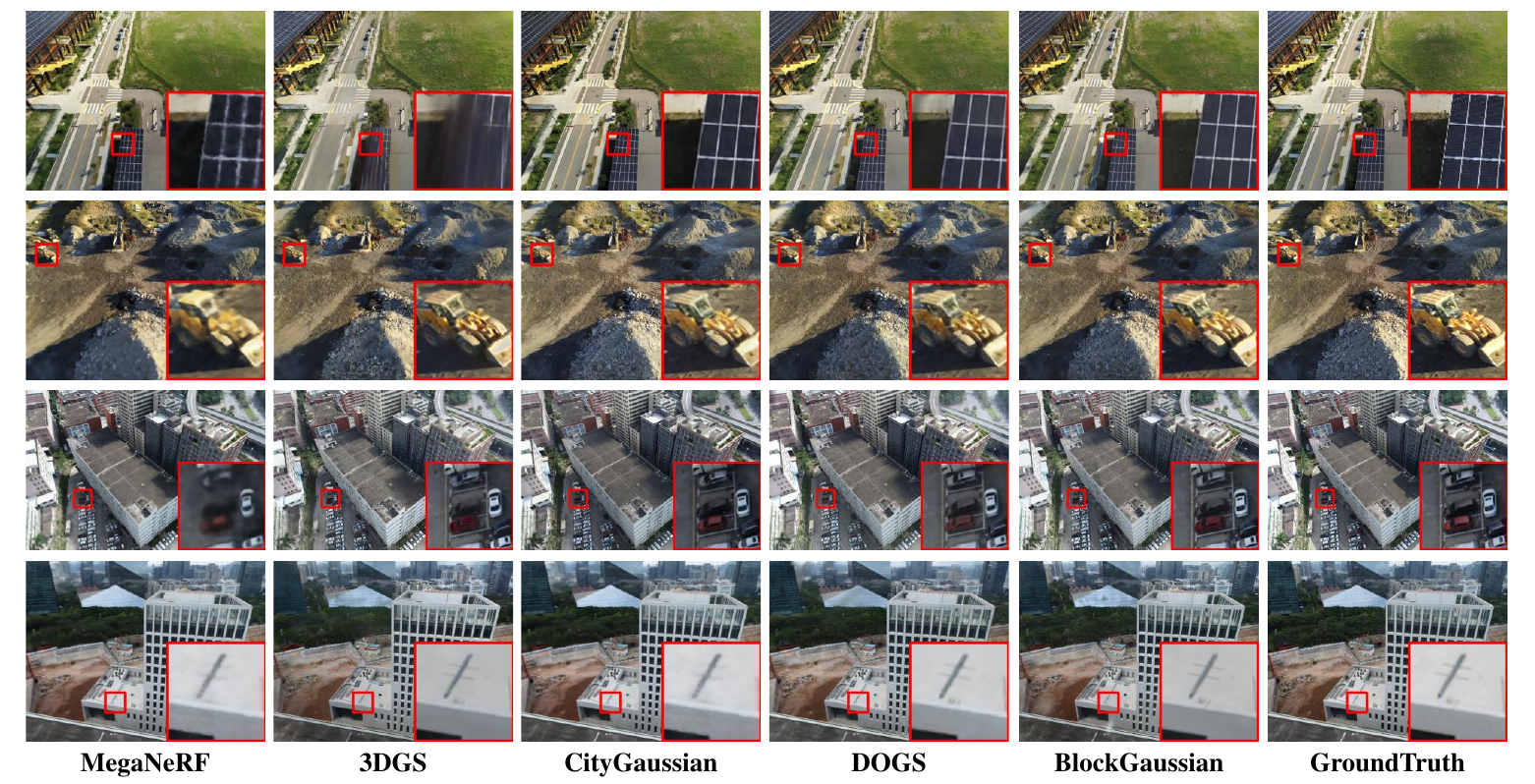}
    \caption{Qualitative Results on Mill19 and UrbanScene3D Datasets.}
    \label{fig:comparison_u3d}
\end{figure*}

\begin{figure*}[t]
    \centering
    \includegraphics[width=\textwidth]{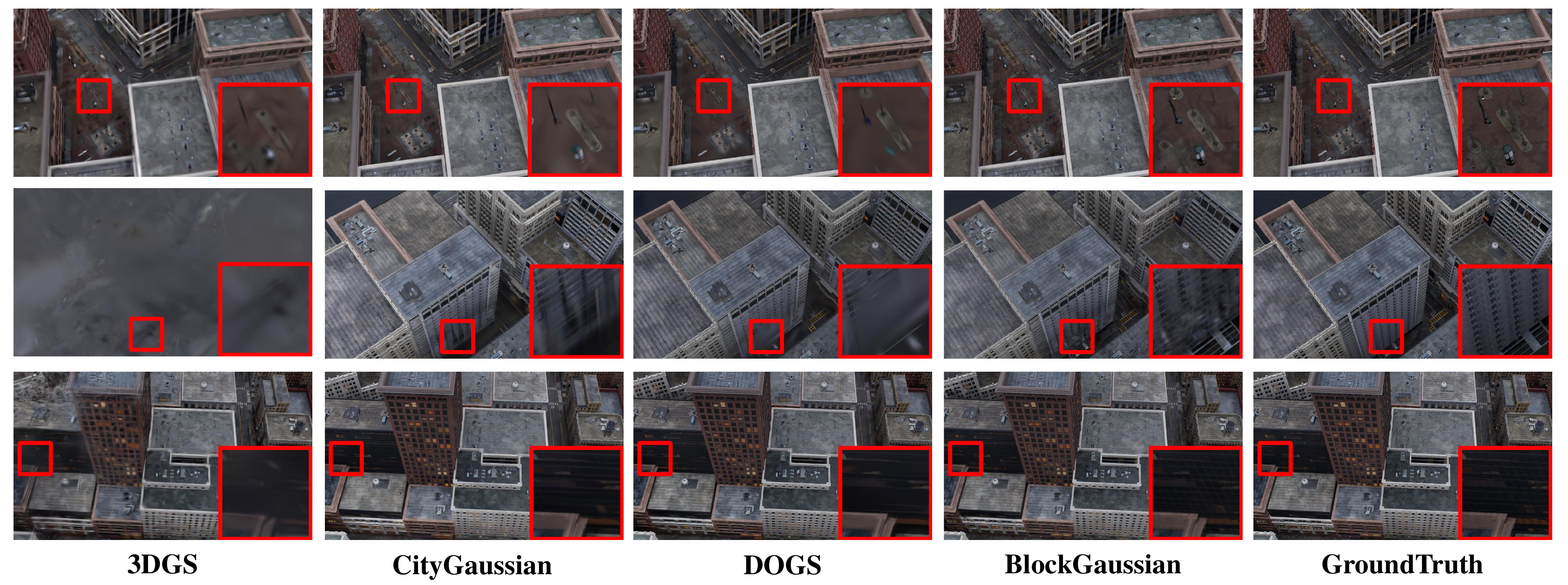}
    \caption{Qualitative Results on MatrixCity Dataset.}
    \label{fig:comparison_mc}
\end{figure*}

\begin{figure*}[t]
    \centering
    \includegraphics[width=\textwidth]{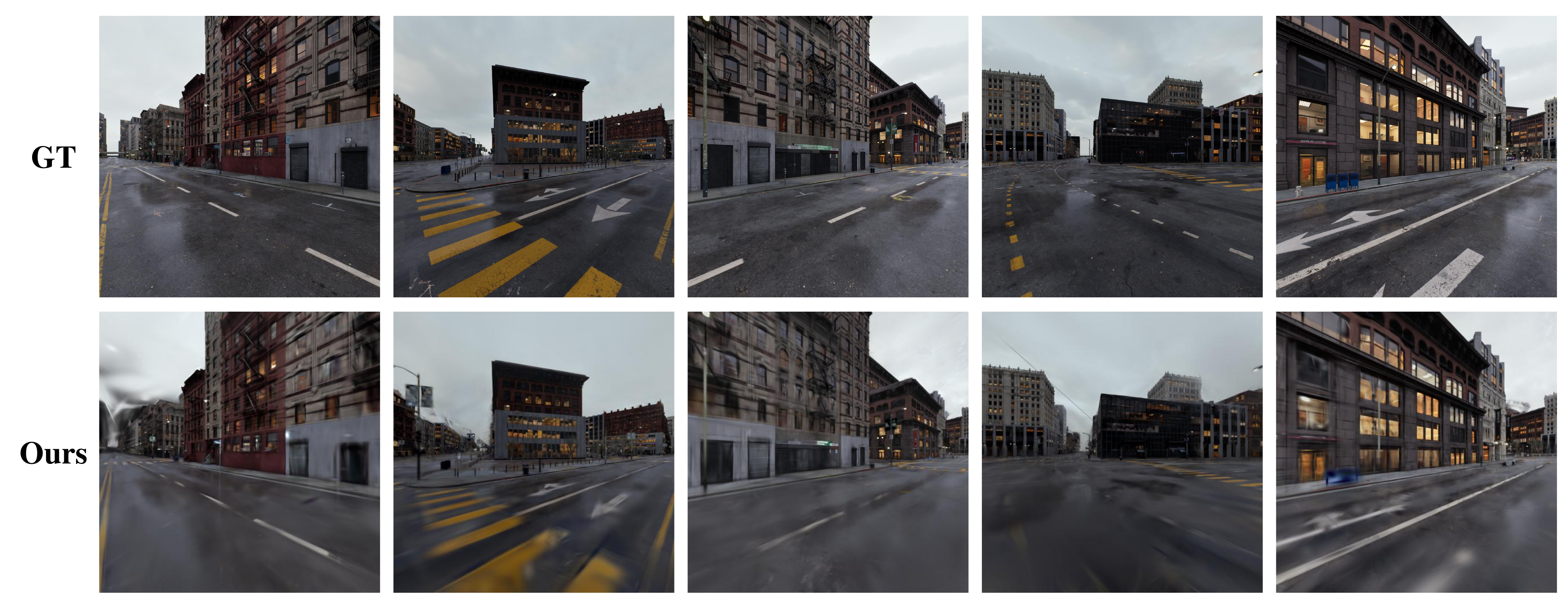}
    \caption{Qualitative Results on MatrixCity Street Scene.}
    \label{fig:mc_street}
\end{figure*}

Compared to NeRF-based methods, BlockGaussian achieves better rendering results. As shown in Fig.\ref{fig:comparison_u3d}, due to implicit scene representation with multilayer perceptron networks, NeRF-based methods tend to produce blurry and overly smooth results. BlockGaussian reconstructs more accurately in high-frequency regions of the scene, which is attributed to point-based representation. Vanilla 3D Gaussian Splatting (3DGS) struggles with areas rich in details (1st row in Fig.\ref{fig:comparison_u3d}) due to insufficient points. The reconstruction results of 3DGS often exhibit numerous floaters in airspace, which are detrimental to interactive rendering. In contrast to Gaussian-based methods, BlockGaussian excels in reconstructing edge and high-frequency regions (1st row in Fig.\ref{fig:comparison_mc}) as well as structurally repetitive areas (2nd row in Fig.\ref{fig:comparison_mc}). In addition to aerial scenes, we also evaluate our method on street-view scene \textit{MatrixCity-Street}. Without any scene-specific tuning, our method demonstrates significant improvements over existing methods, achieving substantial leads in PSNR (+3.87dB), SSIM (+0.169), and LPIPS (-0.377) metrics as shown in Tab.\ref{tab:comparison_mc} and Fig.\ref{fig:mc_street}.

\textbf{Efficiency and Consumption.} As shown in Tab. \ref{tab:comparison_u3d_speed}, we compare optimization time, final point counts, and VRAM consumption across methods. Mega-NeRF, Switch-NeRF, VastGaussian, DOGS, and BlockGaussian are trained on 8 RTX 4090 GPUs, while 3DGS uses a single RTX 4090 GPU. The hyperparameter \textit{Batchsize} of BlockGaussian is set to 1 to match other methods. For CityGaussian, we evaluated metrics using the published checkpoints.
Traditional NeRF-based methods, Mega-NeRF and Switch-NeRF, exhibit significantly longer optimization times (over 19 hours) and require substantial computational resources. In contrast, Gaussian-based approaches demonstrate considerably lower optimization times and more efficient memory usage. Notably, BlockGaussian achieves the fastest optimization times, completing optimization in minutes rather than hours. Increasing the number of optimization iterations from 40K to 60K slightly raises both optimization time and VRAM usage but remains computationally feasible. 
Our method demonstrates significantly faster optimization while generating more points for scene representation, justifying its superior reconstruction quality. This comes at the cost of higher VRAM usage during rendering, but is still much faster than NeRF-based methods.

\subsection{Ablation Study}
We conduct ablation experiments to evaluate the individual contributions of three key components in our proposed framework: Content-Aware Scene Partition, Visibility-aware Block Optimization, and Pseudo-view Geometry Constraint. In addition, we investigate the impact of key hyper-parameters on model performance. The results validate the necessity of each component and provide insights for future improvements and potential simplifications of the framework.

\subsubsection{Content-Aware Scene Partition}
\begin{figure*}[]
    \centering
    \includegraphics[width=\textwidth]{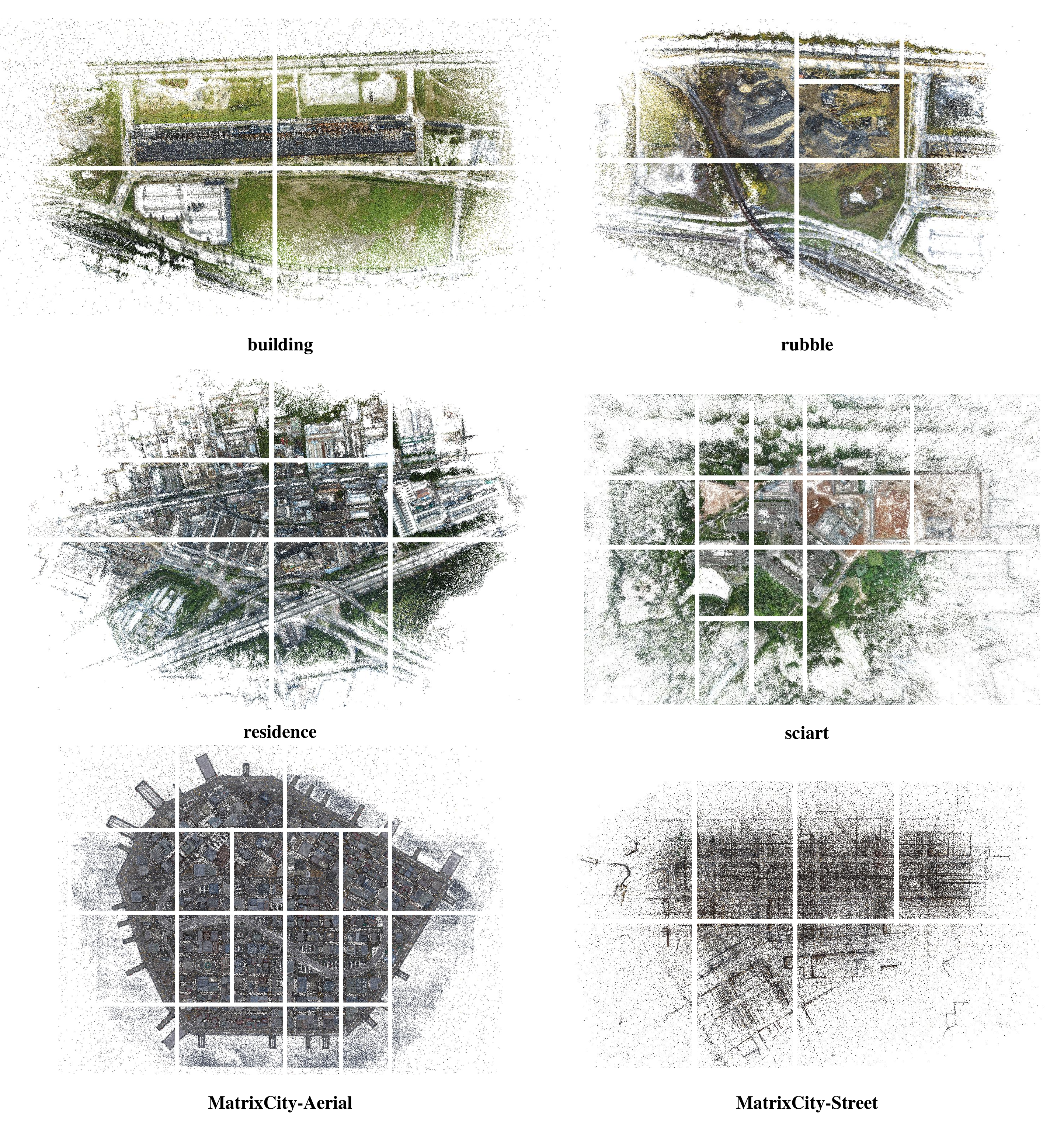}
    \caption{The visualization of scene partition result. The scene is divided into blocks of different sizes according to the density distribution of the sparse point cloud. And the computational load is balanced among multiple blocks.}
    \label{fig:partition}
\end{figure*}

The results of the scene partition are presented in Tab.\ref{tab:scene_partition} and Fig.\ref{fig:partition}. $N_{blocks}$ denotes the total number of blocks partitioned. The terms  $N_{views}^{mean}$ and $N_{views}^{max}$ represent the average and maximum number of views per block, respectively. $N_{pts}^{mean}$ and $N_{pts}^{max}$ indicate the average and maximum number ($10^6$) of initial sparse point clouds within each block. Through Content-Aware Scene Partitioning, we have managed to control the number of initial point clouds within each block to a similar range. This is because the quantity of sparse point clouds roughly reflects the complexity of the scene content in that area. As can be observed from Tab.\ref{tab:scene_partition}, when the number of point clouds within blocks is similar, the variation in the number of views across different scene blocks is considerable, particularly in urban scenes such as \textit{residence} and \textit{MatrixCity-Aerial} scenes. This suggests a weak correlation between the complexity of scene content and the number of views. As illustrated in Fig.\ref{fig:partition}, our proposed strategy enables adaptive scene partitioning based on the distribution of sparse point clouds, thereby balancing the computation complexity across different blocks.

The overall reconstruction speed of the scene is positively correlated with the number of partitioned blocks and the complexity of reconstructing a single block. Since the reconstruction process for each block is entirely independent, block optimization can be performed sequentially on a single GPU or in parallel across multiple GPUs. Here, we report the optimization time of the most time-consuming block denoted as $t_{opt}^{max}$ and the total execution time when processed sequentially on a single GPU denoted as $t_{opt}^{total}$ with the hyper-parameter \textit{Batchsize}=1. The relationship between these times approximately follows $t_{opt}^{total} \approx 0.75 \cdot N_{blocks} \cdot t_{opt}^{max}$, which indicates that the partitioning strategy effectively balances the computational load across all blocks.

\begin{table*}[]
\renewcommand{\arraystretch}{1.3}
\centering
\caption{Scene partition results of multiple datasets.}
\small
\begin{tabular}{cccccccc}
\hline
\textbf{Scenes} & \textbf{$N_{blocks}$} & \textbf{$N_{views}^{mean}$} & \textbf{$N_{views}^{max}$} & \textbf{$N_{pts}^{mean}$} & \textbf{$N_{pts}^{max}$} & \textbf{$t_{opt}^{max}$} & \textbf{$t_{opt}^{total}$} \\ \hline
\textbf{building} & 8 & 660 & 707 & 0.44 & 0.47 & 01:09 & 06:28 \\
\textbf{rubble} & 6 & 434 & 711 & 0.35 & 0.59 & 00:52 & 04:11 \\
\textbf{residence} & 7 & 622 & 1078 & 0.35 & 0.57 & 01:01 & 04:57 \\
\textbf{sciart} & 8 & 342 & 940 & 0.32 & 0.42 & 00:51 & 06:21 \\
\textbf{MC-Street} & 7 & 874 & 1129 & 0.27 & 0.36 & 00:53 & 04:42 \\
\textbf{MC-Aerial} & 16 & 496 & 838 & 0.26 & 0.37 & 00:54 & 10:52\\ \hline
\end{tabular}
\label{tab:scene_partition}
\end{table*}

\begin{figure}[]
    \centering
    \includegraphics[width=0.47\textwidth]{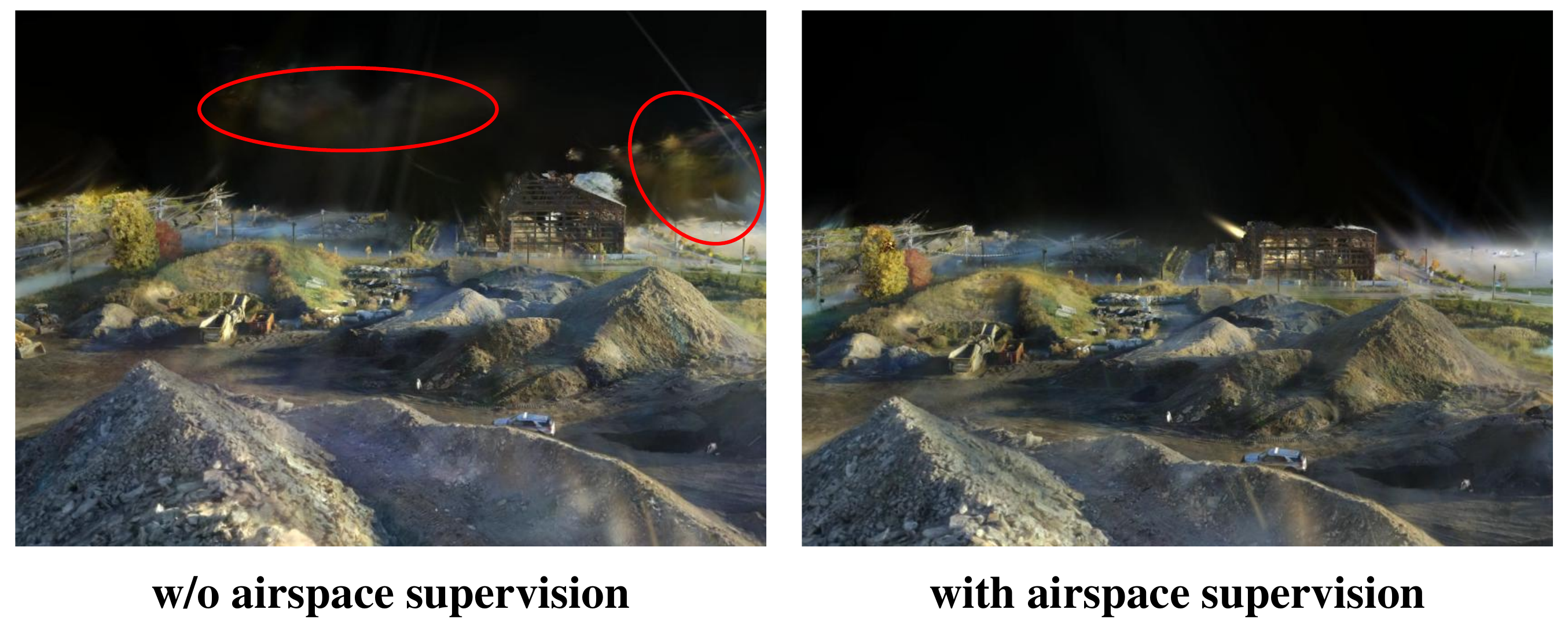}
    \caption{The visualization of Pseudo-View Geometry Constraint ablation experiment. Artifacts are marked within red circles.}
    \label{fig:result_airspace}
\end{figure}

\begin{table}[]
\tiny
\renewcommand{\arraystretch}{1.3}
\centering
\caption{Ablation experiments of our method.}
\resizebox{0.47\textwidth}{!}{
\begin{tabular}{cccccc}
\hline
Aux & $B_{\text{opt}}$ & $L_{\text{pse}}$ & PSNR & SSIM & LPIPS \\ \hline
$\times$      & $\times$       & $\times$     & 24.60  & 0.787 & 0.240 \\
$\times$      & $\checkmark$   & $\times$     & 24.87  & 0.807 & 0.216 \\
$\checkmark$  & $\times$       & $\times$     & 25.45  & 0.809 & 0.213 \\
$\checkmark$  & $\checkmark$   & $\times$     & 26.23  & 0.823 & 0.205 \\
$\checkmark$  & $\checkmark$   & $\checkmark$ & 26.33  & 0.824 & 0.200 \\ \hline
\end{tabular}
}
\label{tab:ablation}
\end{table}

\begin{table}[]
\tiny
\renewcommand{\arraystretch}{1.3}
\centering
\caption{Ablation experiments of the number of blocks.}
\resizebox{0.47\textwidth}{!}{
\begin{tabular}{ccccc}
\hline
$N_{\text{block}}$ & $N_{\text{GPU}}$ &PSNR & SSIM & LPIPS \\ \hline
2 & 2 & 26.99 & 0.829 & 0.199 \\
4 & 4 & 26.09 & 0.827 & 0.199 \\
6 & 6 & 26.33 & 0.824 & 0.200 \\
8 & 8 & 26.16 & 0.819 & 0.207 \\ \hline
\end{tabular}
}
\label{tab:ablation_blocks}
\end{table}

\begin{table}[]
\small
\renewcommand{\arraystretch}{1.35}
\centering
\caption{Ablation experiments of optimization hyper-parameters.}
\resizebox{0.47\textwidth}{!}{
\begin{tabular}{cccccc}
\hline
\textit{Batchsize} & PSNR & SSIM & LPIPS & Points & OptTime \\ \hline
1 & 25.89 & 0.810 & 0.211 & 12.23 & 00:52 \\
2 & 26.12 & 0.816 & 0.208 & 12.76 & 01:30 \\
3 & 26.18 & 0.822 & 0.206 & 12.90 & 02:09 \\
4 & 26.33 & 0.824 & 0.200 & 13.10 & 02:46 \\ \hline
\end{tabular}
}
\label{tab:ablation_minibatch}
\end{table}

\subsubsection{Visibility-aware Block Optimization}

\begin{figure}[t]
    \centering
    \includegraphics[width=0.47\textwidth]{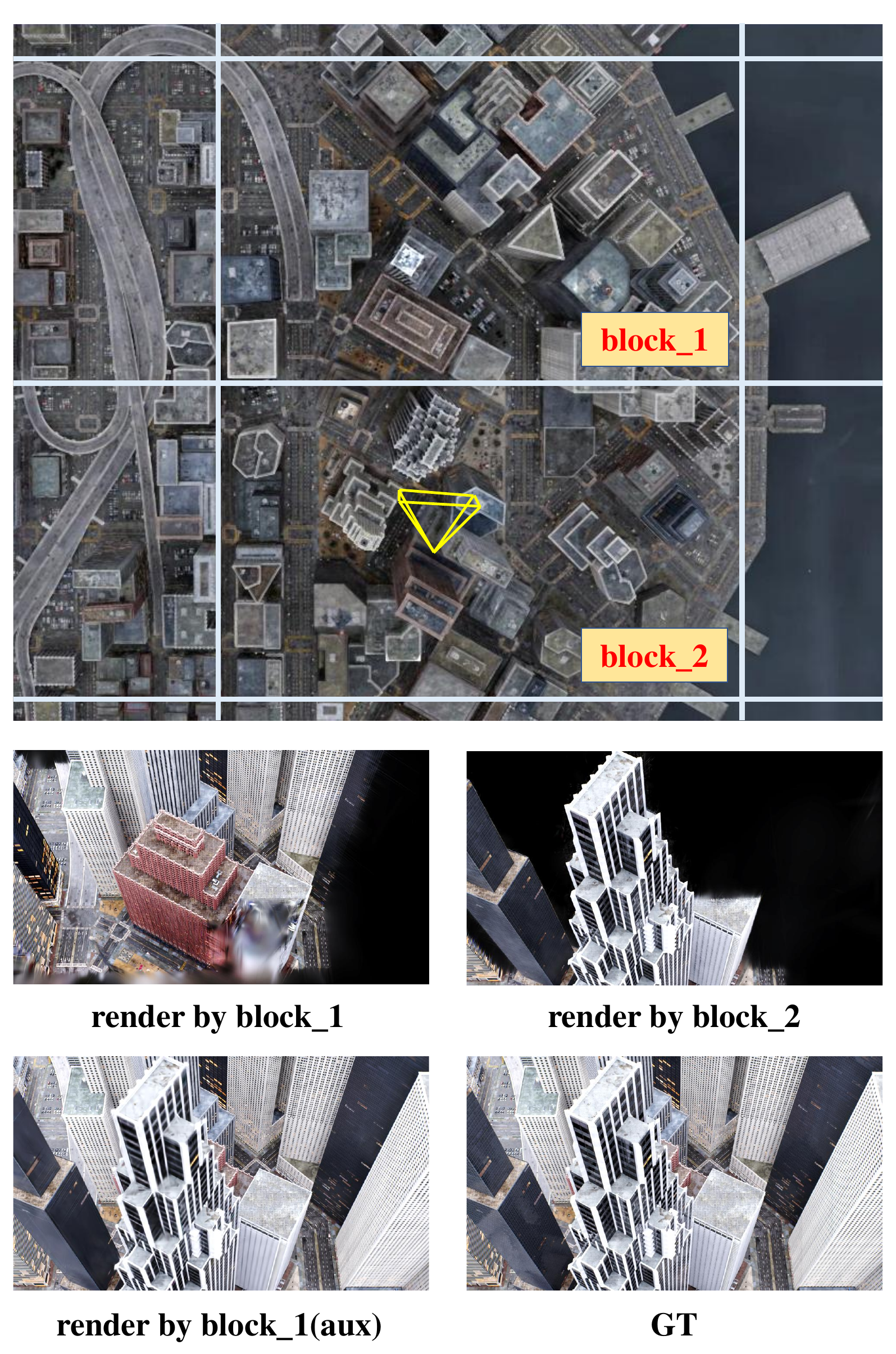}
    \caption{Visualization of block optimization results. Top: local scene partition result. Middle: rendered by points in block. Bottom: rendered by block and auxiliary points, ground-truth image.}
    \label{fig:result_aux}
\end{figure}

As illustrated in Tab.\ref{tab:ablation} and Fig.\ref{fig:result_aux}, we report the ablation study results of Visibility-Aware block Optimization in the \textit{rubble} scene. \textbf{Aux} and $B_{\text{opt}}$ refer to auxiliary points and mini-batch strategy during block optimization. From the 1st and 3rd row of Tab.\ref{tab:ablation}, the quality of scene reconstruction is significantly improved by incorporating auxiliary points. We visualize the rendering results of a supervised view rendering with block points $\mathcal{G}_b$ and auxiliary points $\mathcal{G}_a$. After scene partitioning, the content of the training view is divided into two blocks, and this view simultaneously supervises the reconstruction process of both blocks. When optimizing \textit{block\_1}, the auxiliary points accurately fills the invisible regions of the supervised view (3rd row of Fig.\ref{fig:result_aux}), which indicates that we have achieved an accurate match between the current block and the visible area of the supervised image. Thereby, in the optimization results of \textit{block\_1} and \textit{block\_2}, no floaters are generated (2nd row of Fig.\ref{fig:result_aux}).

Lines 1-2 of Tab.\ref{tab:ablation} demonstrate that the mini-batch optimization strategy improves scene reconstruction quality. In addition, we observe that in scenes with significant lighting variations, optimizing the scene with mini-batch effectively mitigates the generation of floaters in airspace regions, which benefit from more stable gradients during densification. By combining the mini-batch optimization strategy and auxiliary points, our method achieves a notable improvement of +1.6 dB in PSNR, along with corresponding enhancements in SSIM and LPIPS metrics.

\subsubsection{Pseudo-view Geometry Constraint}
As shown in Line 6 of Tab.\ref{tab:ablation}, the Pseudo-view Geometry Constraint contributes to a measurable improvement in the metrics on the test views, indicating that BlockGaussian can reconstruct more accurate and consistent geometry. This effect becomes even more pronounced when wandering through the scene, as illustrated in Fig.\ref{fig:result_airspace}. By supervising the airspace region, the floaters are effectively mitigated, significantly enhancing the image quality of interactive view synthesis.

\subsubsection{Effect of Batchsize}
In Tab. \ref{tab:ablation_minibatch}, we investigate the effect of hyper-parameter \textit{Batchsize} to the reconstruction quality. Increasing the \textit{Batchsize} leads to a steady improvement in PSNR and SSIM while reducing LPIPS, indicating better reconstruction quality. Specifically, PSNR increases from 25.89 to 26.33, SSIM improves from 0.810 to 0.824, and LPIPS decreases from 0.211 to 0.200, demonstrating that larger batch sizes contribute to enhanced perceptual and structural fidelity. This improvement can be attributed to enhanced gradient stability, which effectively facilitates the densification process. Meanwhile, this improvement comes at the cost of increased optimization time. The trade-off between performance and computational cost should be considered when selecting an appropriate batch size for practical applications.

\subsubsection{Number of blocks}
We investigate the effect of number of blocks in BlockGaussian. By adjusting the hyper-parameter maximum tree depth $M$ and block point number threshold $N_{b}^{t}$ in Context-Aware Scene Partition stage, we realize the scene partition of different block numbers.
As shown in Table \ref{tab:ablation_blocks}, increasing the number of blocks brings variation in PSNR metric. Meanwhile, this introduces slight variations in perceptual metrics: SSIM and LPIPS fluctuate within a narrow range. We attribute this PSNR decline to inter-block illumination inconsistencies, where localized lighting variations affect PSNR more significantly than perceptual metrics.

\section{Discussion\label{Discussion}}
Although BlockGaussian demonstrates impressive optimization speed and view synthesis quality, several limitations remain. First, similar to the original 3D Gaussian representation, BlockGaussian requires a substantial number of points to represent intricate scene details. Enhancing the compactness of point cloud representation, as exemplified by LightGaussian\cite{fan2024lightgaussian}, represents a promising direction for improvement. Furthermore, to achieve interactive rendering for large-scale scenes, integrating Level-of-Detail (LoD) techniques\cite{ren2024octree} with dynamic map loading becomes essential. Such integration would enable better compatibility between large-scale scene 3D Gaussian representation and existing rendering pipelines.

\section{Conclusion\label{conclusion}}
This paper introduces BlockGaussian, a framework for novel view synthesis for large-scale scenes. The proposed Content-Aware Scene Partition strategically divides the scene while jointly considering the complexity of scene content and the reconstruction computational loads distribution across blocks. Our Visibility-Aware Block Optimization effectively addresses the challenges posed by invisible regions in supervised views during the reconstruction of individual blocks. The Pseudo-View Geometry Constraint suppresses the generation of floaters in airspace, facilitating the interactive rendering. Notably, our algorithm can be implemented sequentially on a single GPU or in parallel across multiple GPUs. BlockGaussian achieves state-of-the-art performance in view synthesis quality across multiple large-scale scene datasets. In addition, we plan to further explore efficient representations of Gaussian and interactive rendering with dynamic scene map loading in future work.

%  ******************************references section******************************    
\bibliographystyle{IEEEtran}
\bibliography{references}

\end{document}